\documentclass[11pt]{article}

\usepackage{amsmath,amsthm,amssymb,amsfonts}
\usepackage{algpseudocode}
\usepackage{algorithm, mathptmx}
\usepackage{graphicx,cite}
\usepackage{textcomp, fontawesome5}
\usepackage[dvipsnames]{xcolor}
\usepackage[margin=1in]{geometry}
\date{}

\newcommand{\net}{f_{\rm enc}}

\newtheorem{remark}{Remark}

\usepackage{flushend}

\title{\bf Meta-Learning of Neural State-Space Models \\ Using Data From Similar Systems}
\author{Ankush Chakrabarty\thanks{All authors are affiliated with Mitsubishi Electric Research Laboratories, 
			Cambridge, MA 02139, USA. (Corresponding author e-mail: achakrabarty@ieee.org).}, Gordon Wichern, Christopher R. Laughman}

\begin{document}
\maketitle

\begin{abstract}
Deep neural state-space models (SSMs) provide a powerful tool for modeling dynamical systems solely using operational data. Typically, neural SSMs are trained using data collected from the actual system under consideration, despite the likely existence of operational data from similar systems which have previously been deployed in the field. In this paper, we propose the use of model-agnostic meta-learning (MAML) for constructing deep encoder network-based SSMs, by leveraging a combination of archived data from similar systems (used to meta-train offline) and limited data from the actual system (used for rapid online adaptation). We demonstrate using a numerical example that meta-learning can result in more accurate neural SSM models than supervised- or transfer-learning, despite few adaptation steps and limited online data. Additionally, we show that by carefully partitioning and adapting the encoder layers while fixing the state-transition operator, we can achieve comparable performance to MAML while reducing online adaptation complexity.
\end{abstract}


\section{Introduction}
Data-driven system identification is often a necessary step for model-based design of control systems. While many data-driven modeling frameworks have been demonstrated to be effective, the class of models that contain a state-space description at their core have typically been easiest to integrate with model-based control and estimation algorithms, e.g., model predictive control or Kalman filtering. 

Early implementations of neural state-space models (SSMs) employed shallow recurrent layers and were dependent on linearization to obtain linear representations~\cite{zamarreno1998state} or linear-parameter-varying system representations~\cite{bao2020identification}. Recent advancements in deep neural networks have enabled embedding SSMs into the neural architecture explicitly without post-hoc operations~\cite{forgione2022learning}, and therefore the SSM description can be learned directly during training; see~\cite{legaard2021constructing} for a recent survey. For instance, unmodeled dynamics remaining after procuring a physics-informed prior model can be represented using neural SSMs~\cite{forgione2020model,forgione2021dynonet}, and additional control-oriented structure can be embedded during training~\cite{skomski2021automating}. Another interesting direction of research has led to the development of autoencoder-based SSMs, where the neural  architecture comprises an encoder that transforms the ambient state-space to a (usually high-dimensional) latent space, a decoder that inverse-transforms a latent state to the corresponding ambient state, and a linear SSM in the latent space that satisfactorily approximates the system's underlying dynamics~\cite{masti2021learning,iacob2021deep,bertalan2019learning}. Even without the decoder, deep encoder networks have proven useful for neural state-space modeling~\cite{beintema2021nonlinear}. An argument for the effectiveness of autoencoder-based approaches is based on Koopman operator theory~\cite{koopman1932dynamical}, which posits that a nonlinear system (under some mild assumptions) can be lifted to an infinite-dimensional latent space where the state-transition is linear; an autoencoder allows a finite-dimensional, therefore tractable, approximation of the Koopman lifting/lowering transformations~\cite{lusch2018deep}.

Extensively, neural SSMs have been constructed using data from the target system under consideration. However, in practice, one likely has access to data for a range of similar (not necessarily identical) systems that contain information which could prove useful for speeding up the construction of a neural SSM for the target system. Herein, we show that there is potential in using data from similar systems for neural SSM modeling, and provide a meta-learning approach for tractably obtaining such a model. To the best of our knowledge, the existing literature does not contain a tractable solution to this open problem at this time, although the reduction of model estimation error by using data from linear systems within a prescribed ball has been proven in~\cite{9867413}. Though not for SSMs, meta-learning has been proposed for optimization~\cite{zhan2022calibrating,chakrabarty2022meta}, adaptive control~\cite{DBLP:journals/corr/abs-2103-04490}, and receding horizon control~\cite{arcari2020meta,muthirayan2020meta}.

The main \textbf{contribution} of this work is to propose a meta-learning approach for neural state-space modeling, via deep encoder networks, using data obtained from a range of similar systems. In meta-learning, a (often, deep) neural network is trained on a variety of similar ‘source’ system models so that it can make accurate predictions for
a given target system model, with only a few data points from the target system and a small number of gradient-based adaptation steps. Concretely, we learn encoder weights and state/output matrices from a variety of source systems' sensor data. Consequently, even with a small amount of the target system's data, the deep encoder network can be quickly adapted online to obtain a set of encoder weights and state/output matrices representing the target system dynamics. This meta-learned neural SSM is shown via a numerical example to result in higher predictive accuracy than a neural SSM trained solely using target system data, or even a neural SSM trained on the entire source plus target data. We employ model agnostic meta-learning (MAML) algorithms, which are trained by solving a bi-level optimization problem~\cite{finn2017model}, wherein the outer loop extracts task-independent features across a set
of source tasks, and the inner loop adapts to a specific model with a few iterations and limited data. Since the bi-level training paradigm can often lead to numerical instabilities~\cite{antoniou2019train}, a recent variant of MAML, referred to as almost-no-inner-loop (ANIL), slices the network into a base-learner and a meta-learner, and dispenses with (or significantly cuts) inner-loop updates, to improve meta-training performance~\cite{raghu2019rapid}. Another important contribution of this paper is to investigate the trade-off between feature reuse and rapid learning by slicing the deep encoder net into its components and employing ANIL. We show via a numerical example that meta-learning encoder weights while keeping the state/output matrices fixed leads to better predictions than meta-learning state/output matrices while keeping the encoder weights fixed.

The rest of the paper is organized as follows. In Section~\ref{sec:ps} we formally describe the class of systems considered, the model architecture, and the objective of this work. In Section~\ref{sec:maml-deep-ssm}, we describe how the MAML and ANIL algorithms can be used for meta-learning deep neural SSMs. We illustrate the potential of our proposed approach in Section~\ref{sec:results} using a simple nonlinear chaotic system example, and conclude in Section~\ref{sec:conc}.

\section{Preliminaries}\label{sec:ps}

We consider a family of parameterized discrete-time nonlinear systems of the form
\begin{subequations}\label{eq:sys}
	\begin{align}
		x_{t+1} &= f(x_t, \theta_f),\\ 
		y_t &= h(x_t, \theta_h)
	\end{align}
\end{subequations}
where $x_t\in\mathbb R^{n_x}$ denotes the state of the system at time $t\in\mathbb N$ with $x_0$ being the initial state, $y\in\mathbb R^{n_y}$ denotes the measured outputs, $f$ denotes the unknown dynamics, $h$ the unknown output function, and $\theta:=[\theta_f, \theta_h]\in \mathbb{R}^{n_\theta}$ denotes a vector of unknown model parameters.

\begin{figure}[!ht]
	\centering
	\includegraphics[width=\columnwidth]{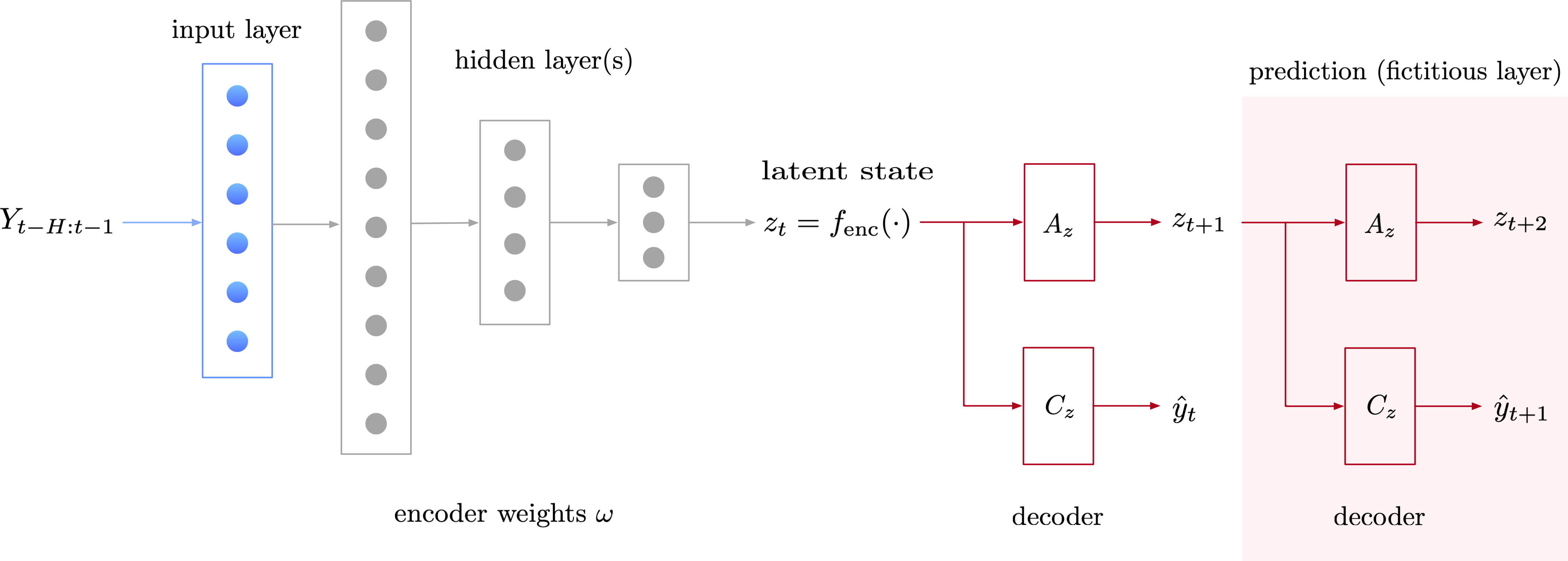}
	\caption{Neural architecture of deep encoder state-space model with an illustration of how to predict using the state-transition operator. The shaded layers are fictitious, and presented for illustrative purposes.}
	\label{fig:enc-ssm}
\end{figure}

Since $f$ and $\theta$ are unknown, our objective is to construct a neural state-space model that can replicate the dynamics of a \underline{query} system of the form~\eqref{eq:sys} parameterized by $\theta^\star$, which is also unknown. Let $Y(\theta^\star, T)$ denote a trajectory of outputs generated by the query system over a time range $[0,T]$, where $T$ is small. This implies that the query system dataset 
\[
\mathcal D_{\rm query} \triangleq Y(\theta^\star, T) \equiv Y^\star
\]
has limited size. One could use this query data to construct a neural state-space model (SSM) of the form
\begin{subequations}\label{eq:deep-ssm}
	\begin{align}
		\label{eq:deep-ssm-a} z_t &= \net(Y_{t-H:t}) \\
		\label{eq:deep-ssm-b} z_{t+1} &= A_z z_t, \\
		\label{eq:deep-ssm-c} \hat y_t &= C_z z_t,
	\end{align}
\end{subequations}
which involves optimizing the weights of an encoder network $\net(\cdot)$, along with the elements of the linear decoders $A_z$ and $C_z$; a schematic diagram of the proposed neural SSM architecture is shown in Fig.~\ref{fig:enc-ssm}. The input to the encoder is a window of length $H\in \mathbb N$ containing past measurements denoted $$Y_{t-H:t}\triangleq \{y_{t-H}, y_{t-H+1}, \cdots, y_{t-1}\},$$ and the latent state learned by the encoder net is given by $z\in\mathbb{R}^{n_z}$, where $n_z\in\mathbb N$ is a hyperparameter. The estimated output is given by $\hat y_t$, which is computed by the decoder $C_z$. 

We reiterate that training the neural SSM involves computing weights of $f_{\rm enc}$ and the linear state-space layers $A_z$, $C_z$. This is performed by minimizing a multi-step prediction loss as follows. For an input $Y_{t-H:t}$, the latent encoding $z_t$ is computed using~\eqref{eq:deep-ssm-a}, after which, for a prediction horizon of $H_{p}$, we recursively compute $z_{t+1},\cdots, z_{t+H_p}$ using~\eqref{eq:deep-ssm-b}. Subsequently, we can compute
\[
\hat Y_{t:t+H_p-1} = \{\hat y_{t}, \hat y_{t+1}, \cdots, \hat y_{t+H_p-1}\}
\]
using~\eqref{eq:deep-ssm-c}. Since training is done offline, the output $y_t$ is available, with which one can construct $Y_{t:t+H_p-1}$. Then, the multi-step predictions can be evaluated via a mean-squared-error (MSE) loss function
\begin{equation}\label{eq:nssm-loss-fn}
	\ell_{\rm SSM} = \tfrac{1}{H_p}\|Y_{t:t+H_p-1} - \hat Y_{t:t+H_p-1}\|_2^2
\end{equation}
that can be minimized using batched data and stochastic gradient descent methods. Often, for better numerical performance, $A_z$ may be regularized with $\mathcal L_1$ (for sparsity) and/or $\mathcal L_2$ (for stability) norms.

\begin{remark}
	Figure~\ref{fig:enc-ssm} illustrates multi-step prediction using a neural SSM with the shaded (fictitious) prediction layer. We emphasize that predicted inputs are not fed back into $\net$, but predicted forward using only linear decoders $A_z$ and $C_z$ without additional inputs. Thus, we can consider $A_z$ and $C_z$ as a type of recurrent neural network, whose initial state is provided by lifting via $\net$. 
	 \hfill$\blacktriangleleft$
\end{remark}

Such neural SSMs have been studied extensively, and various sources have reported their excellent predictive capabilities. However, training the neural SSM using only a limited quantity of target data can lead to poor predictive performance. Our objective in this paper is to use model-agnostic meta-learning (MAML) to learn a generalized neural SSM using data obtained from similar systems to the target system. We can then quickly adapt the generalized neural SSM to the target system despite the scarcity of target data.

To this end, we assume that we have access to a \underline{source} dataset that consists of state trajectories generated by systems of the form~\eqref{eq:sys} parameterized by different $\theta$ vectors, each of which are assumed to be realizations of a distribution $\Theta$ that also admits $\theta^\star$. Then the source dataset is represented by
\[
\mathcal D_{\rm source} = \left\{Y^{k}\right\}_{k=1}^{N_s} \triangleq 
\left\{Y(\theta_k, T_k)\right\}_{k=1}^{N_s},
\]
where $\theta_k\sim\Theta$ for each $k=1,\ldots, N_s$. 

Our \textbf{objective} in this paper is to utilize the source dataset to meta-learn a neural SSM representation offline, and adapt this neural SSM to yield an accurate predictive model for the target system with only a  few online iterations and limited target system data. This is common in practical applications where the amount of data received from a newly deployed target system is typically far less than previously archived customer/user data on similar (source) systems. 

\begin{figure}[!ht]
	\centering    \includegraphics[clip,width=\textwidth]{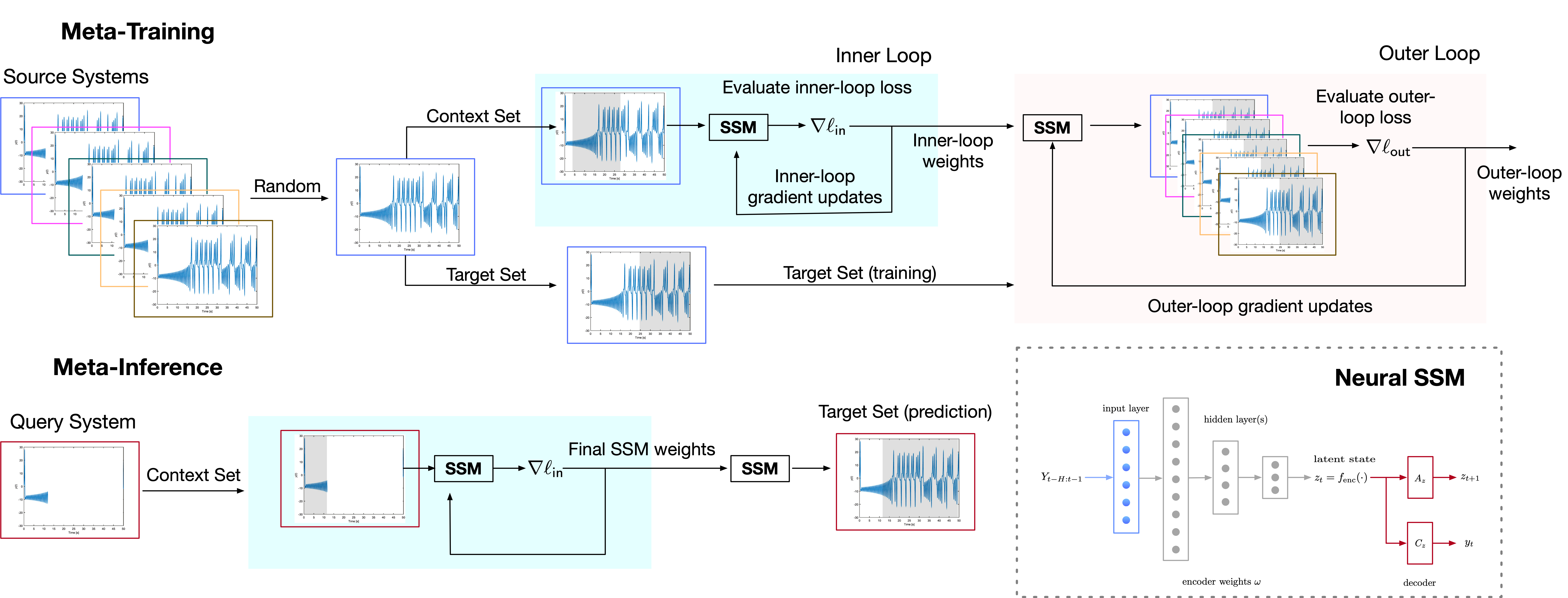}
	\caption{Schematic diagram of the overall meta-learning process.}
	\label{fig:maml-schema}
\end{figure}

\section{Meta-learned Neural SSM}\label{sec:maml-deep-ssm}
\subsection{The MAML Algorithm}


MAML~\cite{finn2017model} is one one of the most well-known and widely used meta-learning (i.e., learning to learn) algorithms. The goal of MAML is to learn a reusable set of model parameters that can be quickly fine-tuned at inference time based on a small amount of adaptation data. MAML achieves this using a nested training scheme where an inner loop fine-tunes a common set of initial model parameters based on a small amount of task-specific adaptation data, and an outer loop that updates the initial set of model parameters across a mini-batch of different tasks. This procedure promotes learning of model parameters that can quickly adapt to new tasks.

An overview of our approach applying MAML for trajectory prediction is shown in Figure~\ref{fig:maml-schema}. Perhaps the most important aspect of applying meta-learning algorithms such as MAML is data partitioning. Formally, we are given a dataset of source systems $\mathcal D_{\rm source}$, comprising trajectory data $Y^i$ for $i=1,\ldots, N_s$ generated by the system~\eqref{eq:sys} with parameters $\theta_i$. We then partition the data for task $Y^i$ into \underline{context} set $\mathcal{C}_{Y^i}$ for inner-loop updates, and \underline{target} set $\mathcal{T}_{Y^i}$ to evaluate the loss function for the model parameters adapted in the inner-loop. If we define $\omega$ as the set of neural network parameters to represent the mapping from~\eqref{eq:deep-ssm}, the inner-loop MAML update is
\begin{equation}\label{eq:maml_inner}
	\omega_{m}^{i}=\omega_{m-1}^{i}-\beta_{\sf in}\nabla_{\omega_{m-1}^{i}}\ell_{\rm SSM}\big(\mathcal{C}_{Y^i};\omega_{m-1}^{i}\big)
\end{equation}
where $m$ is the number of inner-loop updates, $\beta_{\sf in}$ the inner-loop learning rate, and $\ell_{\rm SSM}\big(\mathcal{C}_{Y^i};\omega_{m-1}^{i}\big)$ is the loss function from~\eqref{eq:nssm-loss-fn} evaluated on the context set with the neural weights computed after $m-1$ inner-loop updates. As shown in Figure~\ref{fig:maml-schema}, the inner-loop updates are performed individually using the context set for each trajectory in the batch, while the target sets for all trajectories in the batch are used in the outer-loop. The outer-loop optimization step is typically written as
\begin{equation}\label{eq:maml_outer}
	\omega=\omega-\beta_{\sf out}\nabla_{\omega}\sum_{b=1}^{B}\ell_{\rm SSM}\big(\mathcal{T}_{Y^b};\omega_{m}^{b}\big)
\end{equation}
where $B$ is the number of tasks (i.e., trajectories) in a training mini-batch, $\beta_{\sf out}$ is the outer-loop learning rate, and $\ell\big(\mathcal{T}_{Y^b};\omega_{m}^{b}\big)$ is the loss function on the target set $\mathcal{T}_{Y^b}$ after $m$ inner-loop iterations~\eqref{eq:maml_inner} have been completed. By updating the model parameters in the outer-loop as in~\eqref{eq:maml_outer} across $B$ tasks, we promote a parameter set $\omega$ that can be quickly adapted at inference time. As shown in the bottom of Figure~\ref{fig:maml-schema} at inference-time we only perform $m$ inner-loop updates~\eqref{eq:maml_inner} for a query task $Y^\star$ and evaluate model performance using updated parameters $\omega_m^{\star}$.

\begin{remark}
	When partitioning trajectory data $Y^i$ into context set $\mathcal{C}_{Y^i}$ used for inner-loop fine-tuning and target set $\mathcal{T}_{Y^i}$ used to evaluate the fine-tuning, one may use different approaches for training and inference. At inference time, we will typically use the first several points of the observed trajectory as the context set, and any subsequent points as the target set to simulate the real-world scenario, where we first fine-tune our meta learned model and then use it. At training time, we randomly sample $|\mathcal{C}_{Y^i}|$ consecutive points from anywhere in the trajectory as the context set, and $|\mathcal{T}_{Y^i}|$ consecutive points from anywhere in trajectory (not necessarily after the context set) as the target set. We take this approach during training to ensure that the learned model does not always expect the context set to depend on initial conditions, while target set is in steady-state regions for example. In our experiments we use $|\mathcal{C}_{Y^i}|=|\mathcal{T}_{Y^i}|=12.\hfill\blacktriangleleft$ 
\end{remark}

\subsection{Partial network meta-learning with ANIL}\label{sec:anil}
The nested training scheme in MAML can lead to a very difficult optimization problem~\cite{antoniou2019train}, and updating all model parameters in the inner-loop may be unnecessary~\cite{raghu2019rapid}. For this reason, the almost no inner-loop (ANIL) algorithm was proposed in~\cite{raghu2019rapid}, which updates only the parameters in the last layer of the network in the inner-loop~\eqref{eq:maml_inner}, while parameters for all layers are updated in the outer-loop~\eqref{eq:maml_outer}. The intuition being that the outer-loop update promotes extraction of low-level features that are reusable across tasks, and the inner-loop promotes rapid learning of a final task specific layer. 

In the few-shot image classification problem commonly studied in meta-learning research, the last layer is referred to as a ``classifier'' and the earlier layers that are not updated in the inner-loop as the ``feature extractor.'' However, for the neural SSM architecture studied in this work, it is unclear whether the encoder weights~\eqref{eq:deep-ssm-a} or the state-space model parameters~\eqref{eq:deep-ssm-b} and~\eqref{eq:deep-ssm-c} should be common across tasks or fine-tuned in the inner-loop, so we explore both possibilities. Formally, if $\omega=(\omega_1,...,\omega_L)$, where $\omega_l$ are the network parameters for layer $l$ and $L$ is the total number of layers in the network, we then define $\omega_{\sf in} \subseteq \omega$ as the subset of parameters updated in the inner-loop, and the inner-loop update from~\eqref{eq:maml_inner} becomes
\begin{equation}\label{eq:anil_inner}
	(\omega_l)_{m}^i=
	\begin{cases}
		\omega_l, & \omega_l \notin \omega_{\sf in} \\
		(\omega_l)_{m-1}^{i}-\beta_{\sf in}\nabla\ell_{\rm SSM}\big(\mathcal{C}_{Y^i};(\omega_l)_{m-1}^{i}\big), & \omega_l \in \omega_{\sf in}
	\end{cases}
\end{equation}
The outer-loop of ANIL remains unchanged from MAML, and we note that when $\omega_{\sf in}=\omega$ ANIL and MAML are identical. We summarize the meta-training regime using source tasks via ANIL/MAML in Algorithm~\ref{alg:maml}, and explain how to update the weights for the query system in Algorithm~\ref{alg:maml_inference}. Both algorithms are in the Appendix.

\section{Simulation Results}\label{sec:results}
In this section, we validate our meta-learned neural SSM on a family of parameter-uncertain unforced van der Pol oscillators, where each oscillator is given by
\begin{subequations}\label{eq:vdp}
	\begin{align}
		\dot x_1 &= x_2,\\
		\dot x_2 &= \theta x_2 (1-x_1^2) - x_1,\\
		y &= \begin{bmatrix} x_1 & x_2 \end{bmatrix}^\top.
	\end{align}
\end{subequations}

\subsection{Data collection and implementation details}
The source and target systems are generated by sampling $\theta$ uniformly from the (unknown) range $\Theta=\mathcal U\big([0.5, 2]\big)$, which induces a broad range of dynamics since $\theta$ is effectively a damping parameter. In particular, we simulate until $T=20$ and collect output data for $N_s=200$ source systems with unique $\theta\sim\Theta$ values for each system, with a sampling period of $0.01$~s. The target system $\theta^\star=1.572$, which is also unknown. For each of the source systems, the initial state is randomly sampled from $[-1, 1]^2$, while the target system initial state is fixed at $[1, -0.5]^\top$; the final time is sampled randomly from $T\sim \mathcal U([10, 40])$~s, with the sampling period kept constant for all systems. The past window length $H=10$ and predictive window length $H_p=5$. 

The deep encoder network (see Fig.~\ref{fig:enc-ssm}) we use as the neural SSM has an encoder consisting of an input layer that takes $x$, and passes it through 5 hidden layers, each of which has 128 neurons, to an output layer that generates a latent variable of dimension $n_z=128$. This latent variable is updated using a $128\times 128$ linear layer with no bias, and the output is computed using a $128\times 2$ linear layer with no bias. The entire network uses rectified linear units (ReLUs) for activation, and the weights are initialized using Xavier initialization. The deep neural network is implemented entirely in \texttt{PyTorch}~\cite{pytorch}. For MAML and ANIL implementation, we adapt the open-source PyTorch-based \texttt{learn2learn} toolbox~\cite{learn2learn}. The batch size $B=32$, with the number of inner-loop iterations fixed at $m=10$, and the total number of meta-training epochs is set to $10^4$. The step-sizes for MAML are $\beta_{\sf in}=0.01$ and $\beta_{\sf out}=0.001$. Online, the MAML is allowed $40$ adaptation steps.

\begin{figure}[!ht]
	\centering
	\includegraphics[width=\columnwidth]{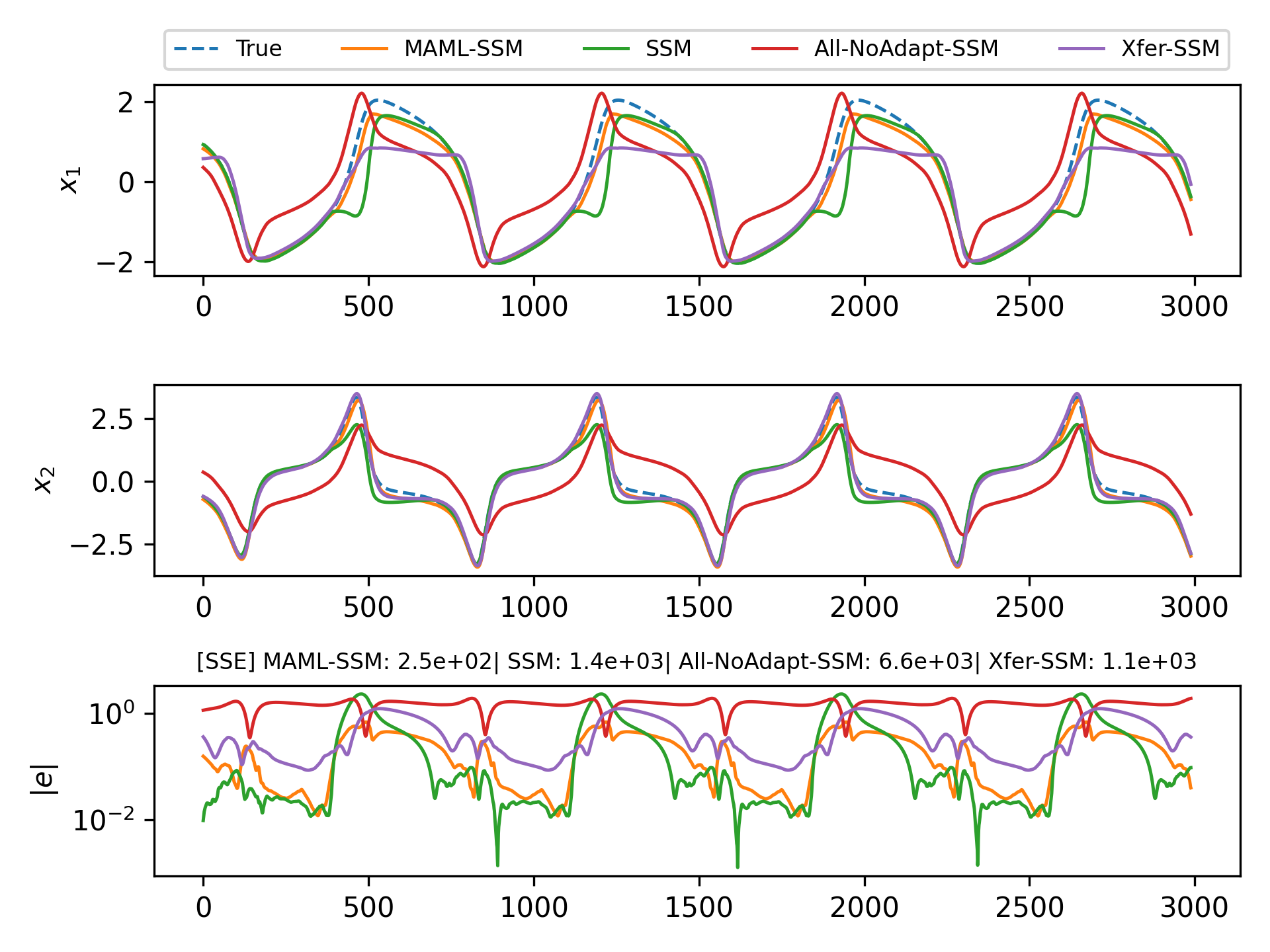}
	\caption{Comparison of MAML-SSM with baselines. (\textit{upper, middle}) State $x_1$ and $x_2$ of the oscillator. (\textit{lower}) Comparison of sum-squared-error (SSE).}
	\label{fig:baselines}
\end{figure}
\subsection{Meta-learning vs. supervised and transfer-learning}

To judge the effectiveness of the proposed \textsf{MAML-SSM}, we compare against a few baselines, all of which have the same architecture as MAML-SSM but are trained differently, using a single training loop rather than an inner-outer bilevel loop as in MAML. These include:
\begin{itemize}
	\item \textsf{SSM}: which is trained by supervised learning using only the target system data: this is the classical one-training-loop approach for learning neural SSMs. The \textsf{SSM} is allowed 10$\times$ more steps to train  
	\item \textsf{All-NoAdapt-SSM}: which is training by supervised learning using all the source system and target system data, without allowing online adaptation steps.
	\item \textsf{Xfer-SSM}, which is trained by transfer learning; that is, supervised learning on all the source system data, and allowing a few online adaptation steps exploiting the target system data. The same number of adaptation steps are allowed for \textsf{MAML-SSM} and \textsf{Xfer-SSM}.
\end{itemize}

To provide some intuition, the first baseline \textsf{SSM} is selected to understand whether \textsf{MAML-SSM} can enable learning from similar systems, or whether using data from other systems negatively affects predictive quality. The second baseline \textsf{All-NoAdapt-SSM} is selected to study whether supervised learning would be enough to learn a good predictive model if it was provided all the data available, both from source and target systems. The third baseline \textsf{Xfer-SSM} is to test whether meta-learning can outperform transfer-learning in few-shot, data-poor situations.

The performance of \textsf{MAML-SSM} and the baselines is illustrated in Fig.~\ref{fig:baselines}. At online inference, the context set is generated by using data from the first 400 time steps, and the target set to be predicted by the neural SSM is the next 3000 time steps. The top and middle subplots show the evolution of the states of the system, with the dashed line denoting the true query system states. The lowest subplot shows the evolution of the sum-squared-error\footnote{The sum-squared-error (SSE) between two signals $\chi, \chi'\in\mathbb R^{d\times T}$ is given by $\sum_{t=1}^T \|\chi_t - \chi_t'\|_2^2$.} between the true state and the predicted state over the 3000 time steps. It is immediately clear that \textsf{MAML-SSM} incurs the smallest prediction SSE and outperforms the baseline, with the transfer learning \textsf{Xfer-SSM} performing second best with double the SSE, closely followed by \textsf{SSM}, and with greedily using all the data as in \textsf{All-NoAdapt-SSM} exhibiting significantly worse performance uniformly across time. From the $x_1$ subplot, we can reason that \textsf{SSM} most likely overfits the context data, and therefore, it exhibits excellent predictive accuracy in within-training-set points of time, with marked deterioration at out-of-set points. Conversely, \textsf{All-NoAdapt-SSM} severely underfits the data, which is plausible, as it inherently tries to find a set of neural SSM weights to fit the `average' dynamics induced by the source and query systems, rather than adapting to the query system. As expected, meta- and transfer-learning exhibit good performance, but our proposed \textsf{MAML-SSM} outperforms \textsf{Xfer-SSM}. The paucity of query system data and the few number of adaptation iterations require the trained learner (before adaptation) to have a set of neural weights that can rapidly adapt to the query system: since this is explicitly what MAML is trained for, the \textsf{MAML-SSM} has the ability to rapidly learn a good predictive model for the given example. In contrast, the transfer learning approach (before adaptation) is not explicitly trained keeping the adaptation in mind, therefore the set of initial weights is likely tailored to generating good predictions for the family of source systems, and the few-shot nature of the adaptation is not enough to enable \textsf{Xfer-SSM} to learn a model as accurate as \textsf{MAML-SSM}; this is especially notable from the $x_1$ plot, where transfer-learning completely fails to capture the positive half of the oscillatory dynamics.

\subsection{Full vs. partial network meta-learning}
As described in Section~\ref{sec:anil}, ANIL can be used to partially meta-learn neural SSMs by splitting the neural SSM into a base-learner (layers with parameters $\omega_l \notin \omega_{\sf in}$) and a meta-learner (layers with parameters $\omega_l \in \omega_{\sf in}$), wherein only the meta-learner is adapted online; the base-learner is fixed. An interesting question for neural SSMs based on deep encoder networks is whether the base-learner should be the encoder layers or the state and output linear layers, as they have clearly different roles to play in the neural SSM. 
To answer this question, we propose two implementations: \textsf{ANIL-SSM}, where the base-learner is $A_z$, $C_z$, with $f_{\rm enc}$ set to be the meta-learner; and, \textsf{ANIL-SSM-R} which has $f_{\rm enc}$ as the base-learner and meta-learns $A_z$, $C_z$.

\begin{table}[!ht]
	\centering
	\caption{Comparison of SSE. Full (MAML) vs. partial (ANIL) meta-learning of neural SSM.}
	\label{tab:anil}
	\begin{tabular}{c||c|c|c}
		\hline
		\textsc{Experiment} & \multicolumn{3}{c}{\textsc{Median SSE}} \\
		\hline \hline
		Size/Steps & \textsf{MAML-SSM} & \textsf{ANIL-SSM} & \textsf{ANIL-SSM-R} \\
		\hline
		200/10 & $\mathbf{1.4\times 10^3}$  & $4.6\times 10^3$ & $6.8\times 10^3$ \\
		200/40 & $\mathbf{1.6\times 10^3}$  & $2.0\times 10^3$ & $6.9\times 10^3$ \\
		200/100 & $\mathbf{1.1\times 10^3}$  & $2.1\times 10^3$ & $1.0\times 10^4$ \\
		500/10 & $\mathbf{1.3\times 10^3}$ & $1.7\times 10^3$ & $6.3 \times 10^3$ \\
		500/40 & $2.3\times 10^3$ & $\mathbf{1.9\times 10^3}$ & $4.1 \times 10^3$ \\
		500/100 & $1.6\times 10^2$ & $\mathbf{6.0\times 10^1}$ & $1.5 \times 10^3$ \\
		1000/10 & $9.1\times 10^2$ & $\mathbf{8.5\times 10^2}$ & $5.5\times 10^3$	\\
		1000/40 & $2.2\times 10^1$ & $\mathbf{2.2\times 10^1}$ & $2.6\times 10^3$	\\
		1000/100 & $3.4\times 10^0$ & $\mathbf{3.1\times 10^0}$ & $5.6 \times 10^2$ \\
		\hline
	\end{tabular}
\end{table}

Table~\ref{tab:anil} compiles the median SSE values obtained over 100 query runs with \textsf{MAML-SSM}, \textsf{ANIL-SSM}, and \textsf{ANIL-SSM-R} from unique and randomly selected initial conditions. The leftmost column of the table indicates the number of context data points used for inference, and the number of online adaptation steps. It is easy to deduce that for small-size context sets, MAML outperforms both ANIL variants, whereas when the context set is larger, and more adaptation steps are allowed, \textsf{ANIL-SSM} performs slightly better than \textsf{MAML-SSM}. The most marked improvement is in the `500/100' case, indicating that ANIL has the potential to surpass MAML, albeit with more context data and adaptation steps. We see that \textsf{ANIL-SSM-R} is uniformly worse, often by orders of magnitude, than its competitors, indicating that rapidly adapting the encoding/lifting transformation with $A_z$ and $C_z$ being reused online without alteration allows the deep encoder network to be more expressive over a wider range of dynamics than reusing the encoding/lifting transformation. This can be explained by realizing that  $f_{\rm enc}$ is a deep network with nonlinear activation functions rather than a single linear layer, and therefore it can induce a much richer family of transformations than $A_z$ and $C_z$, allowing for greater expressivity in modeling nonlinear dynamics.

\section{Conclusions}\label{sec:conc}
In this paper, we proposed a methodology for designing neural state-space models using  limited data from the query system to be modeled, relying more on data from similar systems. We provide a deep learning-based framework called meta-learning for tractably learning from similar systems offline, and adapting to the query system online in a few-shot manner; we also demonstrate the potential  of the proposed approach using a numerical example. Interestingly, we show that meta-learning can outperform transfer learning in data-poor settings when the number of online adaptation steps is limited. We also describe how to  meta-learn a subset of the layers of a neural SSM, and report that meta-learning lifting transformations is more advantageous than meta-learning the state-transition operator itself. In future work, we will investigate alterations to the MAML algorithm for faster training convergence, and apply these techniques to the modeling of energy systems.

\bibliographystyle{IEEEtran}
\bibliography{references.bib}

\appendix

\section*{Pseudocode}
\begin{algorithm}
	\caption{Meta-training SSM with MAML/ANIL}\label{alg:maml}
	\begin{algorithmic}[1]
		\Require $\omega\leftarrow $ weights of neural SSM
		\Require $\mathcal D_{\rm source}\leftarrow$ source task dataset
		\Require $\omega_{\sf in}\subseteq \omega$ \Comment{ANIL: $\omega_{\sf in}\subset \omega$, MAML: $\omega_{\sf in}= \omega$}
		\Require $\beta_{\sf in}$, $\beta_{\sf out}$, $M$ \Comment{learning rates and \# iters}
		\State Randomly initialize $\omega$
		\While{not done} \Comment{outer-loop}
		\State Sample batch $\{Y^b\}_{k=1}^B$ from $\mathcal D_{\rm source}$
		\For{$b=1$ to $B$} \Comment{inner-loop}
		\State Partition $Y^b$ into $\mathcal{C}_{Y^b}$ and $\mathcal{T}_{Y^b}$
		\State $\omega_0^b\leftarrow\omega$ \Comment{copy current weights}
		\For{$m=1$ to $M$} \Comment{adaptation steps}
		\For{$\omega_l \in \omega_{\sf in}$} \Comment{step through layers}
		\State $(\omega_l)^b_m \leftarrow$ update using~\eqref{eq:anil_inner}
		\EndFor
		\EndFor
		\EndFor
		\State $\omega\leftarrow$ update using~\eqref{eq:maml_outer}
		\EndWhile
		\State Return $\omega_\infty\leftarrow$ final trained weights
	\end{algorithmic}
\end{algorithm}

\begin{algorithm}
	\caption{SSM inference with MAML/ANIL}\label{alg:maml_inference}
	\begin{algorithmic}[1]
		\Require $\omega_\infty\leftarrow $ weights of meta-trained neural SSM
		\Require $\mathcal D_{\rm query}\leftarrow$ query task dataset
		\Require $\omega_{\sf in}\subseteq \omega$ \Comment{ANIL: $\omega_{\sf in}\subset \omega$, MAML: $\omega_{\sf in}= \omega$}
		\Require $\beta_{\sf in}$, $M$ \Comment{learning rates and \# iters} 
		\State $\mathcal C_{Y^\star}\leftarrow$ all available data in $\mathcal D_{\rm query}$ 
		\State $\omega_0^\star\leftarrow\omega_\infty$ \Comment{use meta-trained weights}
		\For{$m=1$ to $M$} \Comment{online adaptation}
		\For{$\omega_l \in \omega_{\sf in}$} 
		\State $(\omega_l)^\star_m \leftarrow$ update using~\eqref{eq:anil_inner}
		\EndFor
		\EndFor
		\State Use $\omega_M^\star$ for SSM predictions
	\end{algorithmic}
\end{algorithm}
\end{document}